\PassOptionsToPackage{usenames,dvipsnames,table}{xcolor}
\pdfoutput=1
\documentclass[11pt]{article}

\usepackage[]{acl}

\usepackage{times}
\usepackage{latexsym}
\usepackage{graphicx}
\usepackage{float}
\usepackage{placeins}
\usepackage{multirow}
\usepackage{changepage}
\usepackage{xcolor}
\usepackage{varwidth}
\usepackage{amsmath}
\usepackage{authblk}
\usepackage{hyperref}

\usepackage[T1]{fontenc}
\usepackage[utf8]{inputenc}
\usepackage{booktabs}
\usepackage{microtype}
\usepackage{etoolbox,xspace}

\usepackage{array}
\usepackage{multirow, makecell}
\usepackage[normalem]{ulem} % [normalem] prevents the package from changing the default behavior of `\emph` to underline.
\usepackage{tabularray}

\newcommand{\method}{TaDDEx\xspace}
\newcommand{\dataset}{SymDef\xspace}

\newcolumntype{P}[1]{>{\centering\arraybackslash}p{#1}}

\newtoggle{draft}
\toggletrue{draft}
\togglefalse{draft}
\iftoggle{draft}{
\newcommand{\anna}[1]{\textcolor{magenta}{[#1 \textsc{--anna}]}}
\newcommand{\dk}[1]{\textcolor{cyan}{[#1 \textsc{--dk}]}}
\newcommand{\kl}[1]{\textcolor{red}{[#1 \textsc{--kl}]}}

}{
\newcommand{\anna}[1]{} 
\newcommand{\dk}[1]{}
\newcommand{\kl}[1]{}
}

\newcommand{\com}[1]{}
\title{
Complex Mathematical Symbol Definition Structures: 
\\A Dataset and Model for Coordination Resolution in Definition Extraction}

\author{\textbf{Anna Martin-Boyle$^1$} \ \ \textbf{Andrew Head$^2$} \ \  \textbf{Kyle Lo$^3$} \ \ \textbf{Risham Sidhu$^4$} \ \ \textbf{Marti A. Hearst$^5$} \ \  \textbf{Dongyeop Kang$^1$} \\ 
$^1$University of Minnesota, Minneapolis, MN \ \ $^2$University of Pennsylvania, Philadelphia, PA \\
$^3$Allen Institute for AI, Seattle, WA \ \ $^4$University of Illinois Urbana-Champaign, IL \\
$^5$University of California, Berkeley, CA \\
\tt $^1$\{mart5877,dongyeop\}@umn.edu \ \ $^2$head@seas.upenn.edu \\ \tt $^3$kylel@allenai.org \ \  $^4$rsidhu@illinois.edu \ \ $^5$hearst@berkeley.edu
}

\begin{document}
\maketitle
\begin{abstract}

Mathematical symbol definition extraction is important for improving scholarly reading interfaces and scholarly information extraction (IE). However, the task poses several challenges: math symbols are difficult to process as they are not composed of natural language morphemes; and scholarly papers often contain sentences that require resolving complex coordinate structures. We present \dataset, an English language dataset of 5,927 sentences from full-text scientific papers where each sentence is annotated with all mathematical symbols linked with their corresponding definitions. This dataset focuses specifically on complex coordination structures such as ``respectively'' constructions, which often contain overlapping definition spans. We also introduce a new definition extraction method that masks mathematical symbols, creates a copy of each sentence for each symbol, specifies a target symbol, and predicts its corresponding definition spans using slot filling. Our experiments show that our definition extraction model significantly outperforms RoBERTa and other strong IE baseline systems by 10.9 points with a macro F1 score of 84.82. 
% The competing IE systems DyGIE++, HEDDEx, and SciERC scored 73.92, 64.13, and 63.22, respectively. 
With our dataset and model, we can detect complex definitions in scholarly documents to make scientific writing more readable.\footnote{Our code and dataset are publicly available at \url{https://github.com/minnesotanlp/taddex}}%\footnote{\url{https://github.com/minnesotanlp/taddex} }
\end{abstract}

\section{Introduction}
\begin{figure}[H]
    \centering
    \includegraphics[width=0.4\textwidth]{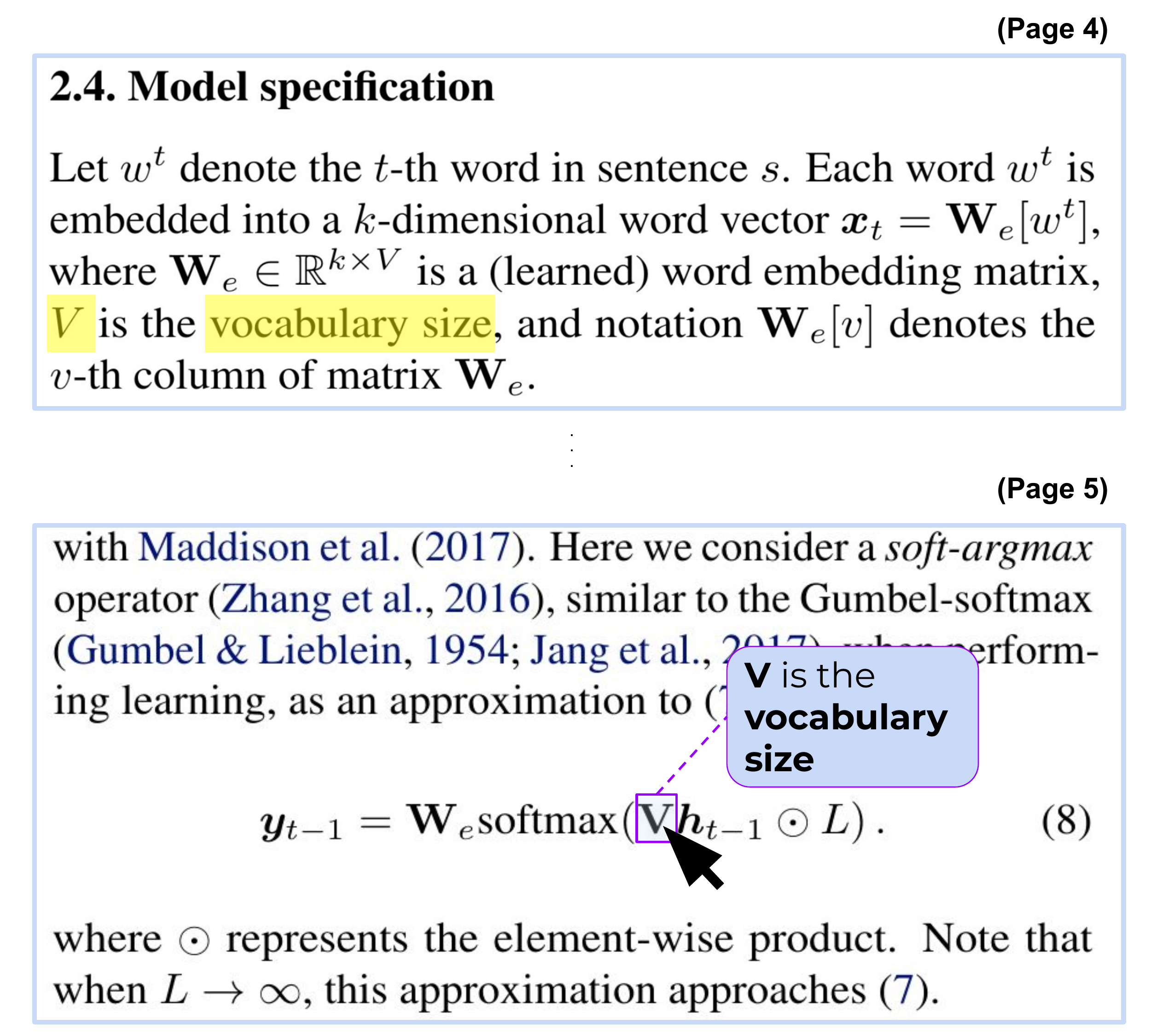}
    \caption{Reading interfaces such as ScholarPhi \citep{Head2021AugmentingSP}  could use math symbol definition extractionto surface symbol definitions as needed. This would save the reader from having to flip between paper sections to look up the definitions of terms in mathematical expressions and algorithms, as in this example from \citet{gu-etal-2018-multimodal}. 
}
    \label{fig:illustration}
\end{figure}

As the volume of scientific publishing increases, it is becoming crucial to develop more sophisticated analysis tools and user interfaces for helping scientists make sense of this ever-growing bounty of knowledge. One particular concern is the ability to accurately extract definitions for mathematical symbols. See Figure \ref{fig:illustration} for one potential use case for mathematical symbol extraction. We find mathematical symbol definition extraction crucial enough to warrant corpora and models tailored to this specific problem.
For definition recognition to be used in user-facing applications, it must achieve a high precision that has not yet been seen in work to date. This task is complicated by the fact that scientific papers often contain multiple symbols and definitions in one sentence, and their spans may be nested or overlapping. Analysis of these symbols and definitions must be coordinated such that the correct definitions are applied to each symbol. Consider for example, the following sentence fragment:

\begin{quotation}
\dots $\mathbf{A}$, $\mathbf{C}$ and $\boldsymbol{\upsilon}$ denote the within-layer adjacency, between-layer adjacency and the community label matrix, respectively.
\end{quotation}

In this case, we wish to define $\mathbf{A}$ as ``within-layer adjacency matrix'', $\mathbf{C}$ as ``between-layer adjacency matrix'', and $\boldsymbol{\upsilon}$ as ``community label matrix''. 

For human readers, the word ``respectively'' immediately clarifies which definition is associated with each symbol. However, even this simple ``respectively'' construction is not obvious to an NLP algorithm, due to the fact that the definitions for $\mathbf{A}$ and $\mathbf{C}$ are split and overlap with the definition for $\boldsymbol{\upsilon}$. 
Little research has been done on the ``respectively'' construct specifically, but other work has found resolution of coordination to be important for resolving hard NLP problems. An error analysis by \citet{fader-etal-2011-identifying} when working on information extraction found that 52\% of errors were in part due to coordination. Information extraction in biosciences (\citealp{ogren-2010-improving}; \citealp{kolluru-etal-2020-openie6}; \citealp{saha-mausam-2018-open}) builds on this insight by attempting to resolve coordination relations directly. \citet{cohen-etal-2009-high} showed that F-scores for recognition of protein-protein structure could be significantly increased by more accurately recognizing coordination structure (using manual rules, assuming distributed semantics, and using post-processing for specific cases). Furthermore, Systems that rely on token-wise structured prediction techniques such as IOB tagging are insufficient to capture complex coordination patterns due to their inability to accommodate overlapping entities.

In order to address the need for improved coordination resolution in scientific information extraction tasks, we developed \dataset, a corpus of scientific papers with a high frequency of complex coordination patterns.  Annotations within \dataset are comprised of mathematical symbols masked as \textsc{SYMBOL} and their sometimes overlapping definitions. This corpus provides an interesting resource for study of complex coordination problems, not only because it contains a high frequency of coordination patterns, but also because the symbols are masked. Because the representations of each symbol are not differentiated  from one another,   the structure and syntax of the sentences are  challenging to identify.

We achieved strong results on our \dataset dataset using a simple but effective method to find the complex mapping between multiple symbols and definitions. Specifically, we decompose the structured prediction problem into multiple passes of definition recognition, with one pass per symbol.  For instance, our method would target the example sentence three times, once for each symbol in \{$\mathbf{A}$, $\mathbf{C}$, $\boldsymbol{\upsilon}$\}, and return the following symbol and definition pairs: <$\mathbf{A}$, ``within-layer adjacency matrix''>, <$\mathbf{C}$, ``between-layer adjacency matrix''>, and <$\boldsymbol{\upsilon}$, ``community label matrix''>.
Since the model recognizes definitions based on a given target symbol, our model is called a \textit{target-based} model. 

Our contributions are the following:
\begin{itemize}
    \item \dataset: a collection of 5,927 sentences from the full texts of 21 scientific papers, with symbols and definitions annotated when present. The papers included contain sentences with complex coordination patterns (such as containing more than two ``and''s or ``or''s, and/or the word ``respectively''). In total, the dataset contains 913 sentences with complex coordination patterns.

    \item The development of a novel target-based approach to definition recognition that isolates one symbol at a time and finds the definition for that symbol within syntactically complex sentences. Our system outperforms two IE baselines and few-shot GPT3 inference by large margins.

\end{itemize}

\section{Related Work}
We discuss previous efforts towards resolving coordination problems, related work in definition recognition, and relevant definition extraction corpora.

\textbf{Syntactic Structure Recognition}.
Coordination is well-studied in linguistics, but analysis is generally in terms of syntactic structure and logical constraints. For example, \citealp{hara-etal-2009-coordinate} focus on creating tree structures or parses for coordination where determining scope is sufficient. Some notable sub-cases such as Argument-Cluster Coordination or right-node raising are often addressed in some way \cite{ficler-goldberg-2016-improved}. There is also work determining the phrase boundaries of components in coordinate structures (\citealp{shimbo-hara-2007-discriminative}; \citealp{ficler-goldberg-2016-coordination}).

While previous work on the syntactic structure of linguistic coordination is useful, definition structures in our work are sometimes more flexible or varied. Furthermore, \citet{dalrymple-1995} found that determining the meaning of ``respectively'' constructions is based on semantics and pragmatics, not the syntax of coordinating conjunctions, and \citet{ogren-2011-thesis} found that a parser-free approach works better than one based on a syntactic parse for interpretation of coordinate structures.

\citet{teranishi-etal-2017-coordination} propose a neural model that uses similarity and substitutability to predict coordinate spans. 
Other work has focused on the problem of splitting sentences into two semantically equivalent ones \cite{ogren-2010-improving}. However, none of the previous work on coordinated definition structures is applied towards using the resolution of coordination patterns for the extraction of term-definition pairs. 

Closest to our work is that of \citet{saha-mausam-2018-open} which splits conjunctions to form multiple coherent simple sentences before extracting relation tuples. One constraint is that multiple coordination structures in a sentence must either be disjoint or completely nested, which is more restrictive than our approach. 

\textbf{Definition Recognition}.
We have found that the ``respectively'' construct is frequently used in the definition of mathematical terms, but its use is not discussed in the literature on definition detection. Others have noted the importance of complex conjunctions in biomedical texts: \citet{ogren-2010-improving} notes that there are 50\% more conjunctions in biomedical scientific text than in newswire text, and \citet{tateisi-etal-2008-genia} also found that coordinating conjunctions occur nearly twice as often in biomedical abstracts as in newswire text. This greater frequency of complex conjunctions in scientific and biomedical texts is significant, as \citet{saha-etal-2017-bootstrapping} found that coordination was the leading cause of IE recall errors.

Also relevant to our work is that of \citet{dai-2018-recognizing}, who summarized the state of the art in discontiguous span recognition, and \citet{dai-etal-2020-effective}, who proposed a transition-based model with generic neural encoding for discontinuous named entity recognition, focusing on separated components and overlapping components. 

Span-based information extraction models such a SciIE \citep{luan-etal-2018-multi} and DyGIE++ \citep{wadden-etal-2019-entity} are relevant for the task of extracting overlapping or nested entities in that span-representations are better-suited to capture overlapping tokens than traditional IOB tagging approaches; for this reason, we use SciIE and DyGIE++ as baseline models (see Section \ref{section:experiments}). 
 
\textbf{Related Corpora}.

There are a few related datasets annotated for definition extraction. The word-class lattices (WCL) dataset \citep{navigli-etal-2010-annotated} comprises 4,564 sentences from the Wikipedia corpus, 1,717 of which have a single definition and 2,847 of which contain false definitions (patterns that resemble definitions but do not qualify as such). 
The W00 dataset \citep{jin-etal-2013-mining} contains 2,512 sentences taken from 234 workshop papers from the ACL Anthology, 865 of which contain one or more non-overlapping definitions. 

The Definition Extraction from Texts (DEFT) corpus \citep{spala-etal-2019-deft} was developed with the intention to provide a more robust corpus of definition annotations with a higher incidence of complex data samples than the WCL and W00 datasets. DEFT includes 2,443 sentences from 2017 SEC filings and 21,303 sentences from open source textbooks on a number of subjects, with a total of 11,004 definition annotations. The DEFT corpus accommodates cross-sentence definitions and multiple definitions per sentence, but not overlapping or nested terms and definitions.

Observing that the extraction of definitions from math contexts requires specialized training corpora, the authors of the Wolfram Mathworld (WFM) corpus \citep{vanetik-etal-2020-automated} developed a corpus of full sentence definitions. 
This corpus comprises 1,793 sentences taken from 2,352 articles from Wolfram Mathworld, 811 of which contain a single definition. 

Most similar to our corpus is the NTCIR Math Understanding Subtask corpus \citep{inproceedings-kristianto-2012}. This corpus contains 10 ArXiv papers with annotated math expressions and their descriptions. Similarly to ours, the annotation scheme allows for discontinuous descriptions. The primary difference between \dataset and the NTCIR corpus is \dataset's focus on overlapping definition and respectively cases. The 21 papers in \dataset were specifically selected because they had relatively high counts of the word ``respectively'' and sentences with multiple ``and''s, and our approach accommodates overlapping definitions (see Section \ref{sec:dataset} for details).

\section{\dataset: Coordination Dataset} \label{sec:dataset}
\dataset is annotated for the coordination of mathematical symbols and their definitions in order to provide a resource for training smart reading interfaces to recognize symbol definitions with a high level of precision. The corpus contains 5,927 English language sentences from the full texts of 21 machine learning papers published on arXiv\footnote{\url{https://arxiv.org/} ArXiv's \href{http://www.overleaf.com}{Terms of Use} states that users may ``Retrieve, store, and use the content of arXiv e-prints for your own personal use, or for research purposes''. As such, \dataset should continue to be used in personal or research contexts.}. These papers were selected by ranking arXiv publications from 2012 to 2022 by the number of mathematical symbols and coordination patterns. This ranking was performed by counting qualifying coordination patterns in each paper, where higher coordination pattern counts were prioritized. These counts were determined per paper using regex pattern matching, searching for the strings ``\texttt{respectively}'' and ``\texttt{, and}''. The highest ranked papers were manually inspected and 21 papers were chosen based on prevalence of symbol-definition pairs.

The first round of annotations was performed by a subset of the authors. This round contributed to the majority of the dataset, resulting in the annotation of 5,661 sentences comprising the full texts of 20 papers.

Additional data were created to supplement the train dataset by annotating another paper containing 226 sentences. These annotations were performed by two domain experts hired through Upwork, one holding a PhD and the other a doctoral student, both in mathematics. The annotators were selected from a set of four applicants to the Upwork listing, all of whom reside in the United States. During the screening process, each of the four applicants were provided with training videos and written documentation in advance, and were trained for 10-30 minutes on 10 example sentences. Their example annotations were monitored and they were asked questions about their process. Upwork annotators were compensated during training and during the annotation task with an hourly rate of \$25.00. Each annotator tracked their hours and were paid \$543.75 each for their work. Upwork applicants were informed of the intended use of the data in the job description, and signed an Agreement form.

All annotations were performed using the annotation software BRAT\footnote{\url{https://brat.nlplab.org/}, \href{https://github.com/nlplab/brat/blob/master/LICENSE.md}{view license here}}.

\begin{figure}[t]
    \centering
    \includegraphics[width=0.48\textwidth]{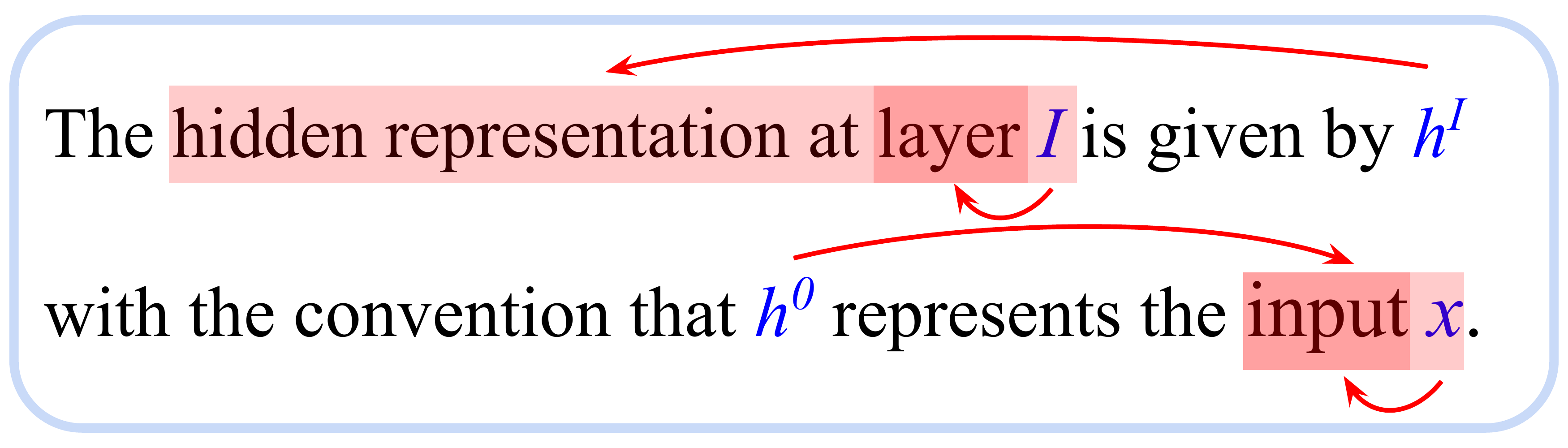}
    \caption{An annotation example for sentences with nested symbols and definitions. $I$ is defined as ``layer'', $h^t$ is defined as ``hidden representation at layer $I$'', $h^0$ is defined as ``input $x$'', and $x$ is defined as ``input''.}
    \label{fig:annotation-example}
\end{figure}

\subsection{Annotation Schema}
The annotation goal for our dataset was to link symbols with the spans of text that define them at the sentence level. In our formulation,  definition spans must declare the meaning of each target symbol; a detailed description of the annotation scheme appears in Appendix \ref{appendix:guidelines}. For example, definition spans may state what the symbol stands for, what a function does, or the datatype it represents. In the case that the symbol represents a collection, the definition may serve to describe the elements contained by the symbol. However, candidate phrases that merely assign a value to the symbol, describe how the symbol is used or computed, or define the symbol with an equation are not valid. Definition spans do not have to contain contiguous text, as they may be split across different parts of the sentence. Furthermore, definitions are permitted to overlap with each other and with symbols as seen in Figure \ref{fig:annotation-example}.

\subsection{Inter-Annotator Agreement}
In Table \ref{tab:my_label}, precision, recall, and F1 scores for exact term and definition matches were calculated to determine the inter-annotator agreement between the Upworks annotators over a subset of 266 sentences. Additionally, the mean percentage of overlapping tokens for definition spans was calculated.  There was significant agreement between annotators for term identification, earning an F1 score of 0.9. Definition identification was more difficult, yielding an F1 score of 0.67 for exact span matches. 
However, on average 85\% of each definition span overlapped between annotators, indicating that, while it is difficult to find the exact span boundaries, annotators were still in agreement on parts of the definition span. 

Of the definition annotations that are not perfect matches, 26 of the annotations from one annotator are contained in the annotations from the other. 126 overlap without containment, with an average number of overlapping words of 4.8. Additionally, 7 of the annotations differ completely, without any overlap.

A review of 1,442 test samples found 76 annotator errors. 46 of these errors were missed definitions. 10 definition spans were nearly correct but contained extra words. 6 were invalid definitions. The remaining errors had to do with improperly defining enumerator variables.

\begin{table}[t]
    \centering
    \begin{tabular}{cccc}
    \toprule
       & \textbf{Term} & \textbf{Definition} & \textbf{Overlap} \\\midrule
        Precision & $.88\pm{.08}$ &  $.65\pm{.12}$ & \multirow{3}*{$85\%\pm{7\%}$}\\\cline{1-3}
        Recall & $.94\pm{.05}$ & $.69\pm{.11}$ & \\\cline{1-3}
        F1 & $.90\pm{.06}$ & $.67\pm{.11}$ & \\
        \bottomrule
    \end{tabular}
    \caption{IAA scores for exact term matches, exact definition matches, and mean percent of definition tokens that overlap in \dataset.}
    \label{tab:my_label}
\end{table}

\subsection{Dataset Characteristics}
We measure the structural complexity of \dataset by considering how many symbols and definitions there are per sentence and how difficult they are to link, and how many sentences contain overlapping or nested symbols and definitions. 

\paragraph{Coordination of Multiple Terms and Definitions}
There are a few characteristics to consider when evaluating the difficulty of coordinating multiple terms and definitions, including: the number of terms and definitions in positive sentences; whether or not every symbol is defined in the sentence (some annotated symbols do not have definitions); and how frequently the terms and definitions are collated (e.g. SYM\dots DEF\dots SYM\dots DEF\dots). The rationale is that an equal number of collated symbols and definitions could be easily coordinated using a simple rule. 

The WCL and WFM corpora contain only one definition per sentence. We compare \dataset with the W00 and DEFT corpora, which sometimes contain multiple terms and definitions per sentence. 

\begin{table*}[h]
    \centering
    \begin{tabular}{c P{1.5cm} P{2cm} P{2.3cm} P{2.3cm} P{1.9cm} P{1.9cm}}
    \toprule
        dataset & \# positive sentences & total terms (terms per sentence) & total defs (defs per sentence) & \# equal term and def. counts & \# collated terms and defs & term | def IAA \\\midrule      
        SymDef & 1,403 & 3,290 \textbf{(2.34)} & 1,713 \textbf{(1.22)} & 681 \textbf{(49\%)} & 576 \textbf{(41\%)} & \textbf{0.90 | 0.67}\\
        W00 & 865 & 959 (1.11) & 908 (1.05) & 725 (84\%) & 699 (81\%) & - | -\\
        DEFT & \textbf{7,311} & 7,847 (1.07) & 7,262 (0.99) & 5,220 (72\%) & 6,582 (90\%) & 0.80 | 0.54
        \\
        %NCTIR\* & - & 2,805 & 1,532 & - & - & - | -
        \\\bottomrule
    \end{tabular}
    \caption{Column 2 shows the total number of sentences containing at least one term. 
    Column 4 shows the total number of definitions. 
    Columns 5 and 6 show the number of samples containing an equal number of terms and with collated terms and definitions. 
    Column 7 shows the reported Inter-Annotator Agreement scores (DEFT was evaluated using Krippendorf's alpha). Boldface indicates the best value per column.
    }
    \label{tab:counts2}
\end{table*}

\paragraph{Overlapping Symbols and Definitions}
\dataset is uniquely focused on the problem of overlapping symbols and definitions, containing 179 sentences with overlapping spans (13\% of positive sentences). Furthermore, many sentences with overlap contained multiple instances of overlapped symbols and definitions. Across all positive sentences there were 480 instances of overlapping, implying that sentences with overlapping contain 2.68 instances on average. W00 and DEFT datasets do not contain overlapping annotations.

\section{\method: Coordination Resolution through \bf{Tar}geted \bf{D}efinition \bf{Ex}traction}

Our aim is to coordinate multiple terms and definitions through targeted definition detection. This is achieved by implementing a target-based definition detection model where the target is one symbol from the sample sentence for which a definition must be recognized. 
\begin{figure}[t!]
    \centering

    \includegraphics[width=.49\textwidth]
    {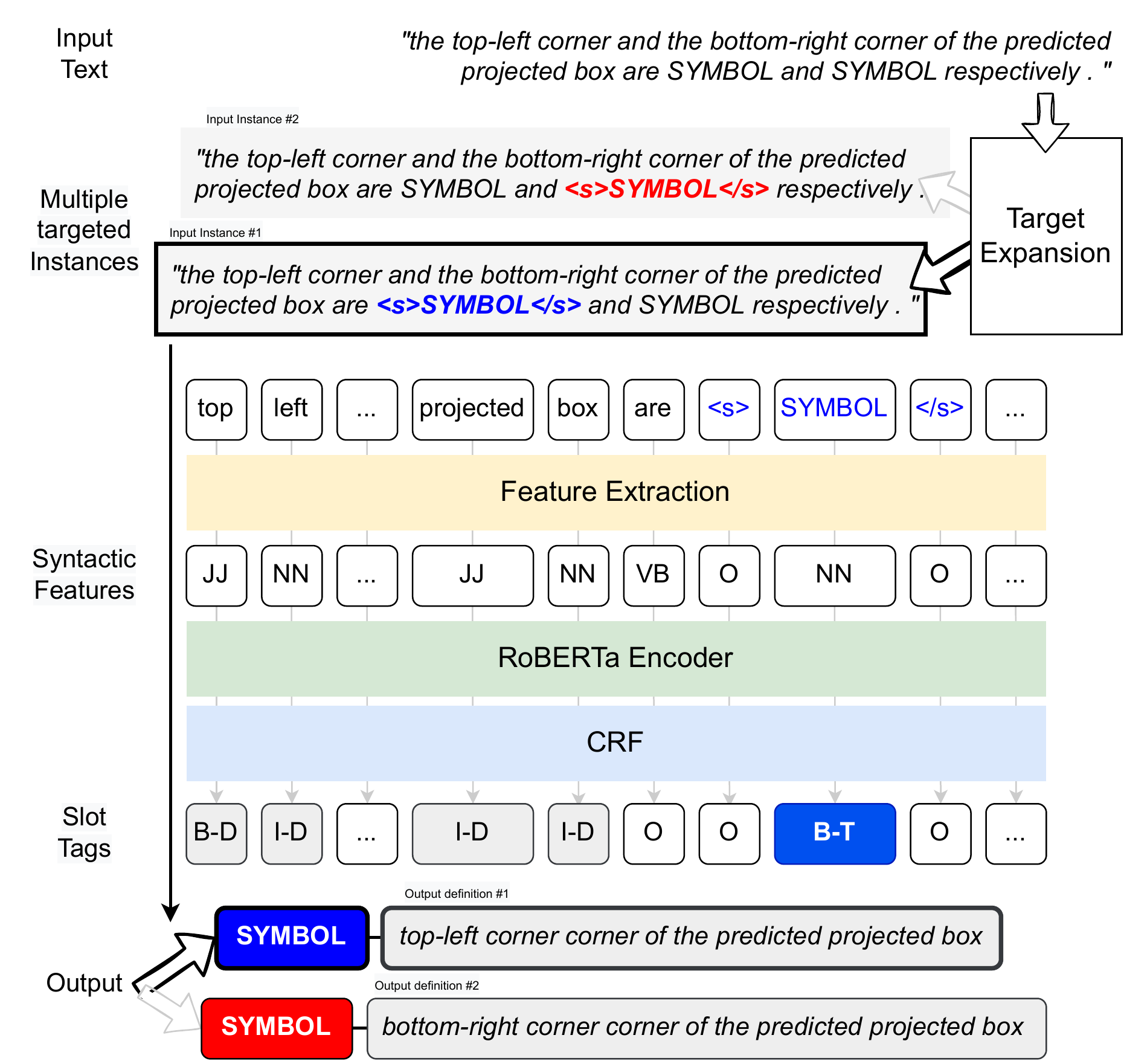}
    \caption{The \method model. A sentence with $n$ symbols is expanded into $n$ samples. Each sample is input into the RoBERTa model individually such that a predicted definition can be recognized for each target symbol.
    }
    \label{fig:my_label}
\end{figure}

\subsection{Targeting Individual Symbols in Complex Coordination}
Mathematical symbols are masked with the term SYMBOL. Sentences with more than one symbol are duplicated once for each additional symbol. For each sample, the symbol for which a definition should be found is tagged as ``</s>SYMBOL</s>''. In this way, each sentence is queried once for each mathematical symbol it contains. For example, the following sentence from \citet{8953532}

\begin{quotation}
\textit{And the top-left corner and the bottom-right corner of the predicted projected box are $( i - S \hat {o} _ { t_{i,j} } , j - S \hat {o} _ { l_{i,j} } )$ and $( i + S \hat {o} _ { b_{i,j} } , j + S \hat {o} _ { r_{i,j} } ] )$ respectively. }
\end{quotation}

would be split into the following two sentences:
\begin{quotation}
\textit{And the top-left corner and the bottom-right corner of the predicted projected box are \textcolor{blue}{\textbf{</s>SYMBOL</s>}} and SYMBOL respectively .}
\end{quotation}
\begin{quotation}
\textit{And the top-left corner and the bottom-right corner of the predicted projected box are SYMBOL and \textcolor{blue}{\textbf{</s>SYMBOL</s>}} respectively . }
\end{quotation}

\subsection{Definition Recognition from Target Symbol}
After an individual symbol is targeted and split into separate instances, we detect a definition of the target symbol.
Our model is built based on the state-of-the-art definition recognition model called Heuristically-Enhanced Deep Definition Extraction (HEDDEx) \cite{kang-etal-2020-document}.
HEDDEx is trained as multi-task learning with two objectives:
it first performs slot-tagging using a Conditional Random Field (CRF)  sequence prediction model.
The model assigns each token in a sentence one
of five tags: term (“B-TERM”, “I-TERM”), definition (“B-DEF”,“I-DEF”), or other (“O”). 
At the same time, a binary classifier is trained to predict a label indicating if the sentence contains a definition.

In detail, after tokenizing the sentences using the ScispaCy\footnote{\url{https://allenai.github.io/scispacy/}, \href{https://www.apache.org/licenses/LICENSE-2.0.html}{Apache License 2.0}} pipeline en\_core\_sci\_md \citep{neumann-etal-2019-scispacy}, we encode input from a Transformer encoder
fine-tuned on the task of definition recognition. 
Following \citet{kang-etal-2020-document}, we choose the best performing Transformer encoder, RoBERTa \citep{Liu2019RoBERTaAR} as our main framework. We used the large version of RoBERTa from Huggingface\footnote{\url{https://huggingface.co/docs/transformers/model_doc/roberta}, \href{https://www.apache.org/licenses/LICENSE-2.0.html}{Apache License 2.0}} \citep{wolf-etal-2020-transformers}. The CRF prediction model we used is torch-crf\footnote{\url{https://github.com/yumoh/torchcrf}, \href{https://mit-license.org/}{MIT Licence}}.

We also provide additional syntactic features as input, which are parts of speech, syntactic dependencies, abbreviations, and entities, which were extracted using ScispaCy.

\section{Experiments}\label{section:experiments}

\paragraph{Datasets}
The dataset is split randomly into train, dev, and test splits. The full texts of papers are kept together for the test set (i.e., sentences in the test set are not members of papers in the train set). The training set contains 4,930 samples after splitting each sentence into samples according to the number of symbols. The dev and test sets contain 1,442 samples each. The data is managed using PyTorch's dataloader\footnote{\url{https://pytorch.org/docs/stable/data.html}, \href{https://github.com/pytorch/pytorch/blob/master/LICENSE}{view license here}}\citep{NEURIPS2019_bdbca288}.

\begin{table}[t!]
    \centering
    \begin{tabular}{@{}c@{\hskip 2mm}c@{\hskip 2mm}c@{\hskip 2mm}c@{\hskip 2mm}c@{}}
    \toprule
    \textbf{Model} & & \textbf{Macro} & \textbf{Term} & \textbf{Def}\\\toprule
    \multirow{3}*{TaDDEx (ours)} & \cellcolor{Gray!50} F & \cellcolor{Gray!50} \textbf{84.82} & \cellcolor{Gray!50} \textbf{81.54} & \cellcolor{Gray!50} \textbf{73.56}\\
    & \cellcolor{red!30}P & \cellcolor{red!30}82.08 & \cellcolor{red!30}74.83 & \cellcolor{red!30}71.91 \\
    & \cellcolor{blue!30}R & \cellcolor{blue!30}\textbf{88.04} & \cellcolor{blue!30}89.56 & \cellcolor{blue!30}\textbf{75.28} \\\midrule
    \multirowcell{3}{HEDDEx \\\citep{kang-etal-2020-document}} & \cellcolor{Gray!50}F & \cellcolor{Gray!50}64.13 & \cellcolor{Gray!50}64.63 & \cellcolor{Gray!50}36.03 \\
    & \cellcolor{red!30}P & \cellcolor{red!30}64.80 & \cellcolor{red!30}61.68 & \cellcolor{red!30}44.37 \\
    & \cellcolor{blue!30}R & \cellcolor{blue!30}64.26 & \cellcolor{blue!30}67.87 & \cellcolor{blue!30}30.33 \\\midrule
    \multirowcell{3}{SciIE \\\citep{luan-etal-2018-multi}} & \cellcolor{Gray!50}F & \cellcolor{Gray!50}63.22 & \cellcolor{Gray!50}53.16 & \cellcolor{Gray!50}37.49\\
    & \cellcolor{red!30}P & \cellcolor{red!30}84.76 & \cellcolor{red!30}79.53 & \cellcolor{red!30}76.47\\
    & \cellcolor{blue!30}R & \cellcolor{blue!30}54.85 & \cellcolor{blue!30}39.92 & \cellcolor{blue!30}24.83\\\midrule
    \multirowcell{3}{DyGIE++ \\ \citep{wadden-etal-2019-entity}} & \cellcolor{Gray!50}F & \cellcolor{Gray!50}73.92 & \cellcolor{Gray!50}65.44 & \cellcolor{Gray!50}57.03\\
    & \cellcolor{red!30}P & \cellcolor{red!30}\textbf{98.02} & \cellcolor{red!30}\textbf{98.41} & \cellcolor{red!30}\textbf{97.05} \\
    & \cellcolor{blue!30}R & \cellcolor{blue!30}63.12 & \cellcolor{blue!30}49.01 & \cellcolor{blue!30}40.38
    \\\midrule
    \multirowcell{3}{GPT3 (few-shot) \\ \citep{NEURIPS2020_1457c0d6}} & \cellcolor{Gray!50}F & \cellcolor{Gray!50}50.51 & \cellcolor{Gray!50}66.30 & \cellcolor{Gray!50}37.22\\
    & \cellcolor{red!30}P & \cellcolor{red!30}43.79 & \cellcolor{red!30}50.53 & \cellcolor{red!30}25.06 \\
    & \cellcolor{blue!30}R & \cellcolor{blue!30}66.53 & \cellcolor{blue!30}\textbf{96.39} & \cellcolor{blue!30}72.31\\
    \bottomrule
\end{tabular}
    \caption{Comparison of definition recognition systems on \dataset: \colorbox{Gray!50}{F1}, \colorbox{red!30}{precision}, and \colorbox{blue!30}{recall} scores. The Macro scores were calculated by finding the mean of the individual scores each of the three labels ``O'', ``I-DEF'', and ``I-TERM''. The Term and Definition scores are a binary measure of the system's ability to classify Terms and Definitions.} 
    \label{tab:results}
\end{table}

\begin{figure*}[t]

\begin{minipage}[[hbt!]{0.65\linewidth}
    \centering
    \small
    \begin{tabular}{P{2cm}@{\hskip 3mm}P{2cm}@{\hskip 12mm}P{2cm}@{\hskip 4mm}P{2cm}@{\hskip 4mm}P{1cm}@{\hskip 4mm}P{1cm}@{\hskip 4mm}P{2cm}} 
    \toprule
        \multicolumn{7}{p{15cm}}{``Each word $w^t$ is embedded into a $k$-dimensional word vector $x_t=W_e\lbrack w^t\rbrack$, where $W_e\in R^{kxV}$ is a (learned) word embedding matrix, $V$ is the vocabulary size, and notation $W_e\lbrack v\rbrack$ denotes the $v$-th column of matrix $W_e$.''}
        \\ \toprule
        \textbf{Term} & \textbf{Gold} & \textbf{TaDDEx} & \textbf{HEDDEx} & \textbf{SciIE} & \textbf{DyGIE++} & \textbf{GPT3}
        \\ \toprule
        $w^t$ & word &  \textcolor{blue}{\underline{word}} &
     \textcolor{blue}{\underline{word}} & \textcolor{blue}{\underline{word}} & \textcolor{blue}{\underline{word}} & 
     each word 
     \\ \midrule
        $k$ & \begin{tabular}{@{}c@{}}-dimensional\end{tabular} & \textcolor{blue}{\underline{-dimensional}} & \textcolor{blue}{\underline{-dimensional}} & - & - & \textcolor{blue}{\underline{-dimensional}}
        \\ \midrule
        \begin{tabular}{@{}c@{}}
        $x_t=$ $\mathbf{W_e}[w^t]$
        \end{tabular} 
        & \begin{tabular}{@{}c@{}}$k$-dimensional\\word vector\end{tabular} & \begin{tabular}{@{}c@{}}\textcolor{blue}{\underline{$k$-dimensional}} \\ \textcolor{blue}{\underline{word vector}} \end{tabular} & - & - & - & word vector  
        \\\midrule
         \begin{tabular}{@{}c@{}}$\mathbf{W}_e\in$ $ \textbf{R}^{k\times V}$ \end{tabular}
         & \begin{tabular}{@{}c@{}}( learned ) \\ word  \\ embedding \\  matrix\end{tabular} & \begin{tabular}{@{}c@{}}learned ) \\  word  \\ embedding \\  matrix\end{tabular} & \begin{tabular}{@{}c@{}}learned )  \\ word  \\ embedding  \\ matrix\end{tabular} & - & - & \begin{tabular}{@{}c@{}}learned  \\ word  \\ embedding  \\ matrix\end{tabular} 
         \\\midrule
         $V$ & vocabulary size & \textcolor{blue}{\underline{vocabulary}} \textcolor{blue}{\underline{size}} & \textcolor{blue}{\underline{vocabulary}} \textcolor{blue}{\underline{size}} & - & - & \textcolor{blue}{\underline{vocabulary}} \textcolor{blue}{\underline{size}}
         \\\midrule
         $\mathbf{W_e}[v]$  & notation $v$-th column of matrix $\mathbf{W_e}$ & \textcolor{blue}{\underline{notation $v$-th}} \textcolor{blue}{\underline{column of}} \textcolor{blue}{\underline{matrix $\mathbf{W_e}$}} & notation & - & - & notation 
         \\\midrule
         $v$ & - & column & -th column of matrix $\mathbf{W_e}$& - & - & -th column 
         \\\midrule
         $\mathbf{W_e}$ & matrix & \textcolor{blue}{\underline{matrix}} & - & - & - & \textcolor{blue}{\underline{matrix}} 
         \\\bottomrule
    \end{tabular}
\end{minipage}
\caption{An example ground-truth annotation in the test of \dataset: (left) a complex sample including the terms, definitions, and relations between them. (right) Eight ground-truth and predicted term-definition pairs. Exact correct definitions are shown in \textcolor{blue}{\underline{blue}}. Nothing output shown as -. From \citet{10.5555/3305890.3306095}.}\label{fig:complex_example}
\end{figure*}

\paragraph{Baselines}
We trained and tested two span-based information extraction models on our dataset, SciIE \citep{luan-etal-2018-multi} and DyGIE++ \citep{wadden-etal-2019-entity}.  We transformed our dataset into the SciIE format, where TERM and DEF are named entities, and DEFINITION-OF is the relation between coordinated terms and their definitions. Mathematical symbols were masked with SYMBOL, but the models were not pointed towards a targeted symbol. Instead, the models were tasked with extracting multiple TERM and DEFINITION pairs per training sample. Each model's ability to coordinate multiple terms and definitions was measured by looking at its ability to extract DEFINITION-OF relations between the named entities. Details on the  setup for these experiments can be found in Appendix \ref{appendix:reproducibility}.

We also calculated zero-, one-, and few-shot GPT3 baselines using text-davinci-003 in a question-answer format. For details on the experimental setup and post-processing, see Appendix \ref{appendix:gpt3}.

\paragraph{Training}
For \method, we trained RoBERTa large \cite{Liu2019RoBERTaAR} on the tokenized samples and syntactic features from the training set for 50 epochs using a batch size of 12, and maximum sequence length of 100. AdamW\footnote{\url{https://huggingface.co/transformers/v3.0.2/main_classes/optimizer_schedules.html}} is used for optimization, with a learning rate of 2e$-$5 and Adam's epsilon set to 1e$-$6. These hyperparameter settings were based on the results of the parameter sweep performed for \citet{kang-etal-2020-document}. After each epoch, the model is validated on the dev set, and the model weights are updated upon improved performance. Loss is calculated using cross entropy loss\footnote{\url{https://pytorch.org/docs/stable/generated/torch.nn.CrossEntropyLoss.html}}.

\paragraph{Evaluation Metrics}
We used BOI tagging to evaluate model performance, where words in the sample sentence that are not a part of a term or definition are assigned ``O'', terms are assigned ``B-TERM'', and definition spans are indicated with the tags ``B-DEF'' (for the first word in the span) and ``I-DEF'' (for the remaining words in the span). We ultimately merged the ``B-DEF'' and ``I-DEF'' tags. 
The predicted labeling is compared with the ground truth by calculating the macro F1, precision, and recall scores for three classes ``O'', ``B-TERM'', and ``I-DEF''. We also report the F1, precision, and recall scores for ``B-TERM'' and ``I-DEF'' individually. FAll scores were calculated for all models using scikit-learn \citep{scikit-learn}.

\subsection{Main Results}

The evaluation scores can be seen for TaDDEx and the baseline systems in Table \ref{tab:results}. Results were generated with a single run.
Both IE baseline models were able to extract the named entities TERM and DEF, as well as the relation DEFINITION-OF. See Table \ref{tab:results} for the resulting scores.

Figure \ref{fig:complex_example} shows a sample from the test set containing a complicated coordination. This sample has 8 terms and 8 definitions, some of which are overlapping.

\subsection{Error Analysis}
Of the 1,442 test samples, our system made incorrect predictions for 135 samples. Of the 135 errors, 28 ($20.7\%$) of them were false negatives, 33 ($24.4\%$) of them were false positives, and 74 ($54.8\%$) were labeled incorrectly. Often, the system's predicted definition overlapped with the ground truth, but added or omitted tokens. 
Sometimes, the system incorrectly labeled definitions in sentences without a symbol definition.

\begin{figure}[t]
    \centering
    \includegraphics[width=0.50\textwidth,trim={2.9cm 3.5cm 2cm 3.5cm},clip]{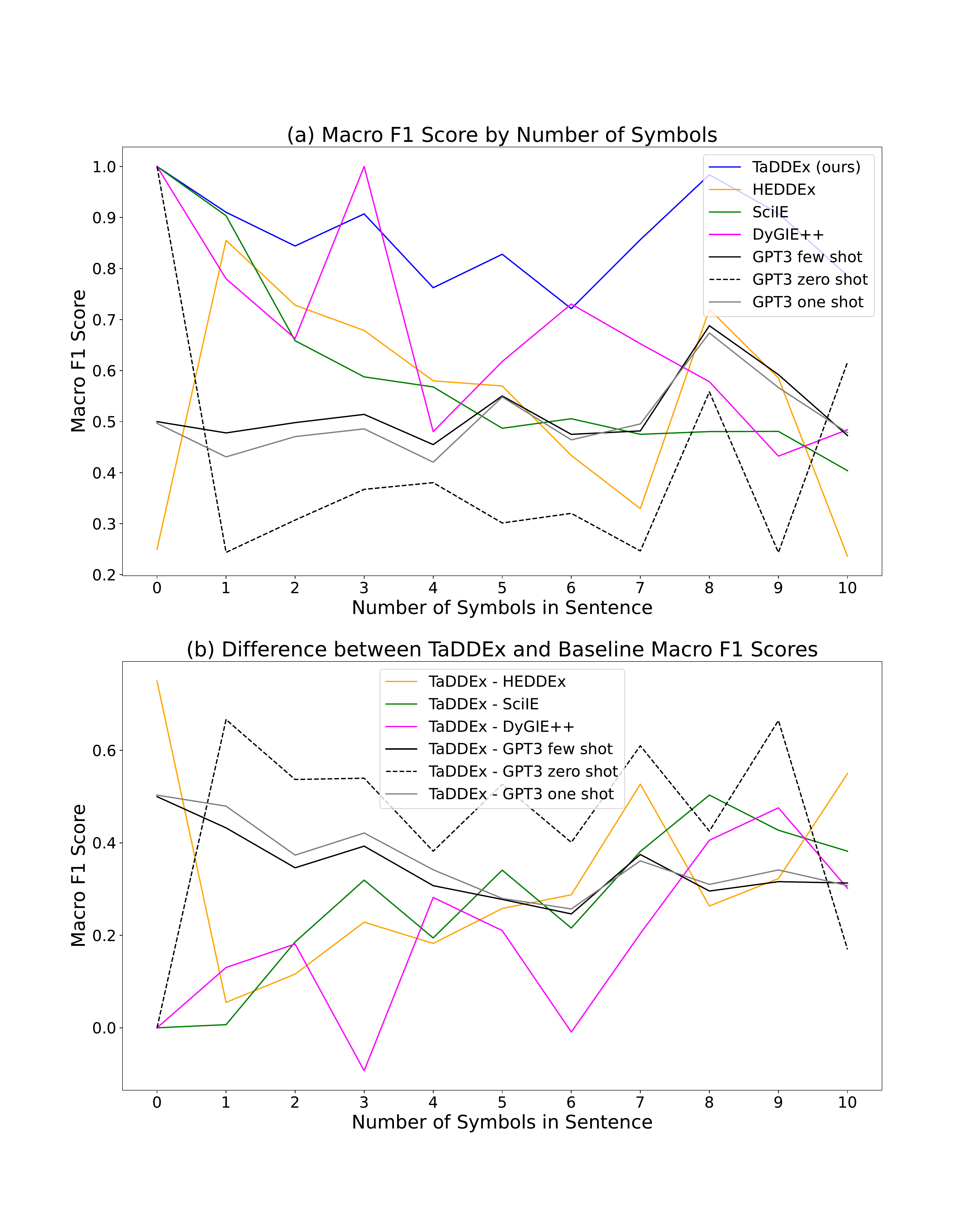}
    \caption{(a) The macro F1 score based on the number of symbols in the sample, and (b) the difference in scores calculated by subtracting baseline F1 scores from TaDDEx.}
    \label{fig:score_per_num_symbol}
\end{figure}

There was not a strong correlation in terms of system accuracy and total number of symbols in the sample for the \method model and GPT3 baselines, but HEDDEx, SciIE, and DyGIE++ performed much better for samples with fewer symbols (see Figure \ref{fig:score_per_num_symbol}). All three systems performed perfectly on sentences without a symbol. TaDDEx was least accurate for sentences with six or ten symbols, but did not generally perform worse as the number of symbols increased: the mean macro F1 score for samples with between 1 and 5 symbols was 85.03 with standard deviation $\pm5.48$, and the mean score for samples with between 6 and 10 symbols was $85.11 \pm9.15$. SciIE's scores decreased as the number of symbols per sample increased from 0 to 5 symbols, remained stable from 5 to 9 symbols (scores ranging between 47.51 and 50.56), then dropped to 40.39 for ten samples. DyGIE++ assigned``O'' to every token, yielding a perfect score for samples with zero symbols, and between 31.84 and 32.84 for all other samples. These results are significant, because they show that the targeted definition recognition method is better at complex term-definition coordination than traditional span-based information extraction techniques.

\section{Limitations and Future Work}
Having to point the model to the term targeted for definition identification requires prior knowledge of the terms in the dataset. This requires either a dataset with annotated terms such as \dataset, or an initial classification step to extract the terms for each sentence.

Within the domain of our \dataset dataset, terms are restricted to mathematical expressions, which are masked with the single token SYMBOL. One limitation of our model is that it under performs for non-symbolic terms. However, we emphasize that the problem of mathematical symbol definition extraction is important enough that it is appropriate to target an approach specifically to this problem. Furthermore, we believe the inability of information extraction systems such as DyGIE++ and SciIE to adapt to the challenges of \dataset warrants the development of approaches that work specifically for the extraction of mathematical symbol definitions.

\section{Potential Risks}
A system that surfaces automatically extracted definitions to readers without 100\% accuracy will occasionally surface an inaccurate definition. Intelligent reading interfaces that use definition extraction run the risk of providing wrong information to the reader. Furthermore, the "illusion of clarity" that information systems provide can elicit a false sense of complete understanding in human users, which discourages the users from looking more closely \citep{nguyen_2021}.

\section{Conclusion}
In this paper we describe the creation of a dataset of 21 scientific papers containing a high concentration of complex term and definition coordinations. We also provide a novel methodology for recognizing multiple coordinated term and definition pairs by targeting one term at a time. Our results show improvement on the span-based approach to relation extraction. Particularly promising is the consistency that our model maintains as the number of symbols per sentence increases.

\newpage
\FloatBarrier
\bibliography{anthology,custom}
% \pagebreak
\newpage
\clearpage
\appendix
\section{Annotation Guidelines}
\label{appendix:guidelines}

The annotation goal was to determine which mathematical symbols in a sentence have definitions; to determine the span of the definitions; and to link the symbols with their definitions. Symbols can take a few forms, including the following:
\begin{itemize}
    \item single letters such as $x$;
    \item composite symbols comprising multiple characters:
    \begin{itemize}
        \item letters with subscripts or superscripts such as $x_i^j$;
        \item function declarations like $f(x,y)$;
        \item and derivative deltas and gradients ($dx$, $\delta x$, $\Delta J$)
    \end{itemize}
    \item and longer patterns such as sequences, expressions, or formulae: 
    \begin{itemize}
        \item $(x_1, x_2, … x_n)$;
        \item  $q_{\phi }(\mathbf {z}|\mathbf {x}) = \mathcal {N}(\mathbf {z}; \mu , \sigma ^2\mathbf {I})$.
    \end{itemize}
\end{itemize}

A definition is a span of text that declares the meaning of the symbol, beginning and ending on word boundaries. Definitions may provide clarity by showing what a symbol represents, the type of information the symbol represents, what a function does, the elements in a collection represented by a symbol, or what differentiates the symbol from other symbols in the sentence. To help identify definitions, the annotators were asked to mark a span of text as a definition if it answers at least one of the questions in Table \ref{tab:is-a-symbol}.

\begin{table*}[]
    \centering
    \begin{tabular}{p{5.3cm}||p{3.8cm}|p{1.5cm}|p{3cm}}
        Question & Sentence & Symbol & Definition \\\hline\hline
        What does this symbol stand for? & ``\dots the function $f$\dots'' & $f$ & ``function'' \\\hline
        What does this function do? & ``$func(x)$ maps a vector to a continuous value.'' & $func(x)$ & ``maps a vector to a continuous value'' \\\hline
        \multirow{2}{5.3cm}{What is the information or type of the data this symbol represents?} & ``\dots the vector $x$\dots'' & $x$ & ``vector'' \\\cline{2-4}
        &``$p$ is a set of programs.'' & $p$ & ``set of programs'' \\\hline
        What are the elements that make up a vector or a set or other collection represented by the symbol? & ``$\Theta$ contains all parameters of the model'' & $\Theta$ & ``contains all parameters of the model'' \\\hline
        What differentiates this symbol from other symbols if there are other related symbols? & ``This produces output embeddings $E_O$ from input embeddings $E_I$.'' & $E_O$ & ``output embeddings''
    \end{tabular}
    \caption{Questions that help determine whether a candidate definition span is valid.}
    \label{tab:is-a-symbol}
\end{table*}

The following constructs may resemble a symbol definition, but did not count as such for this annotation project. 
\begin{itemize}
    \item Equations defining a symbol: ``We define $x$ to be $x=a^2+c$.''
    \item Values assigned to the symbol: ``We set $x$ to 5.''
    \item How the symbol is meant to be used: ``$x$ is then passed as an argument to function $func(x)$ to compute a score.''
    \item How the symbol is computed: ``$x$ is derived by taking the weighted sum of input values.''
    \item The syntactic structure of a phrase implies a meaning without explicitly stating the meaning: ``\dots $i^{th}$ item\dots'' implies that $i$ is an index, but is not explicit so the symbol $i$ does not have a definition. 
\end{itemize}
Additionally, symbols appearing in a label macro or in a standard math operator such as ``log'' or ``sqrt'' should not be annotated. 

We provided instructions on how to determine the boundaries of definition spans. In particular, we specified what kinds of words to include in spans, what kinds of words to omit, and how to determine definition spans when the definition contains non-contiguous tokens. See Table \ref{tab:include-omit} for examples. 
\begin{table*}[]
    \centering
    \begin{tabular}{p{5cm}||p{3.6cm}|p{1.2cm}|p{3.6cm}}
        Guidelines for determining Span Boundaries & Sentence & Symbol & Correct Definition Span \\\hline\hline
        Include multiple definition spans if the definition information is split on either side of the symbol. & ``The function $f$ computes an accuracy score.'' & $f$ & ``function'', ``computes an accuracy score'' \\\hline
        Include multiple definition spans if there are multiple definitions offering distinct interpretations of the same symbol. & ``$f$, the output function, is a linear model.'' & $f$ & ``output function'', ``linear model'' \\\hline
        Include definitions even if they look vague. & ``\dots function $f$\dots'' & $f$ & ``function'' \\\hline
        Include parentheticals that appear within an otherwise contiguous definition span. & ``$f$ is a neural network (NN) for labeling inputs.'' & $f$ & ``a neural network (NN) for labeling inputs'' \\\hline
        Include citations that appear within an otherwise contiguous definition span. & ``$f$ is a spectral neural network CITATION for labeling inputs.'' & $f$ & ``a spectral neural network CITATION for labeling inputs'' \\\hline
        For composite symbols, include definitions of the subsymbols that are part of the composite symbol. & ``$x_i$ is an element at index $i$.'' & $x_i$ & ``an element at index $i$''\\\hline\hline
        Omit determiners (``the'', ``a'', ``some'', etc.) & ``The function $f$\dots'' & $f$ & ``function'' \\\hline
        Omit definition verbs (``is a'', ``means'', ``denotes'', etc.)  & ``$f$ is a function.'' & $f$ & ``function'' \\\hline
        Omit information about the dimensionality or length of data & ``$A$ is a 3x3 array'' & $A$ & ``array'' \\\hline\hline
        \multirow{2}{5cm}{Split definition spans for symbols coordinated with a conjunction} & \multirow{2}{5cm}{``$x$ and $y$ are the model's input and output.''} & $x$ & ``model's'', ``input'' \\\cline{3-4}
        & & y & ``model's'', ``output''
    \end{tabular}
    \caption{Guidelines for what to include and what to omit from definitions}
    \label{tab:include-omit}
\end{table*}

\begin{figure}
    \centering
    \includegraphics[width=0.5\textwidth]{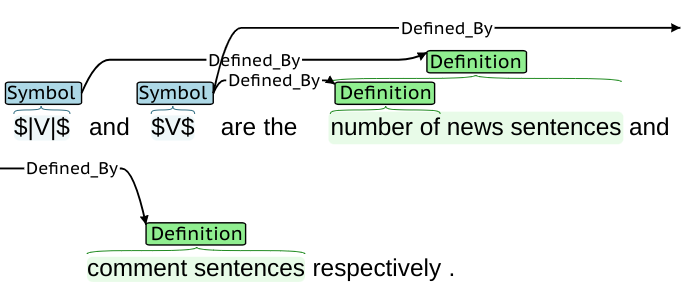}
    \caption{A simple example with the keyword ``respectively''.}
    \label{fig:simple}
\end{figure}

\begin{figure}
    \centering
    \includegraphics[width=0.5\textwidth]{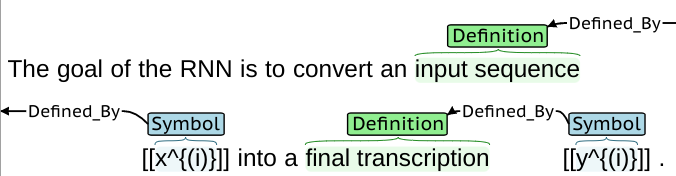}
    \caption{The symbols in this sentence are defined using the appositive structure, where the adjacent nouns ``sequence'' and ``transcription'' define them.}
    \label{fig:appositive}
\end{figure}

\begin{figure}
    \centering
    \includegraphics[width=0.5\textwidth]{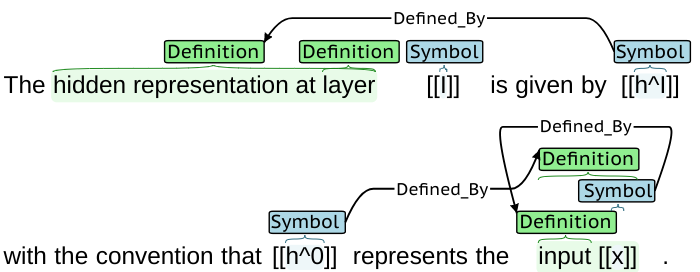}
    \caption{An example of overlapping definitions and symbols. ``layer'' defines ``l'', and ``hidden representation at layer [[l]]'' defines ``$h^l$''. ``Input'' is the definition for ``x'', and ``input[[x]]'' defines ``$h^0$''.}
    \label{fig:overlapping}
\end{figure}

\FloatBarrier
\section{Experimental Setup}
\label{appendix:reproducibility}

\begin{table}[h]
    \centering
    \begin{tabular}{c|c}
        Hyperparameter & Value \\\hline
        epochs & 50 \\
        batch size & 12 \\
        sequence length & 100 \\
        learning rate & $2e-5$ \\
        Adam's epsilon & $1e-6$ \\
        optimizer & AdamW \\
        loss function & cross entropy loss
    \end{tabular}
    \caption{TaDDEx and HeDDEx hyperparameters}
    \label{tab:hyperparams}
\end{table}

\subsection{SciIE Setup}
To reproduce our SciIE experiments, follow these steps:
\begin{enumerate}
    \item Clone or download the \href{https://bitbucket.org/luanyi/scierc/src/master/generate_elmo.py}{scierc repository}.
    \item Create the following directory in the scierc folder: ``./data/processed\_data/json/''. Find our dataset in the SciERC format at \url{anonymous} and copy the train, dev, and test json files to this new directory.
    \item Follow the steps in the README under the ``Requirements'', ``Getting Started'', and ``Setting up for ELMo'' headers. 
    \item Edit the file ``experiments.conf'': find the experiment called ``scientific\_best\_relation'' (at the bottom of the file). Set the coref\_weight to 0 and the ner\_weight to 1.
    \item When running ``singleton.py'' and ``evaluator.py'', pass ``scientific\_best\_relation'' as a command line argument.
    \item Proceed to follow the instructions under ``Training Instructions'' and ``Other Quirks''.
    \item To compare results to TaDDEx, use the script \url{anonymous} to convert the output into our format.
\end{enumerate}

\subsection{DyGIE++ Setup}
To reproduce our DyGIE++ experiments, follow these steps:
\begin{enumerate}
    \item Clone or download the \href{https://github.com/dwadden/dygiepp}{dygiepp repository}.
    \item Create a folder in the repository called ``data/''. Find our dataset in the SciERC format at \url{anonymous} and copy the train, dev, and test json files to this new directory.
    \item Setup your environment with the requirements specified in the README.
    \item Navigate to the ``training\_config'' folder and copy ``scierc.jsonnet'' to a new file called ``symdef.jsonnet''. 
    \item Open ``symdef.jsonnet'' and update ``data\_paths'' so that ``train'' is set to ``data/train.json'', ``validation'' is set to ``data/dev.json'', and ``test'' is set to ``data/test.json''.
    \item Run ``bash scripts/train.sh symdef'' to train the model. To evaluate, follow the instructions in the README under the header ``Evaluating a model''. You will run a command such as ``allennlp evaluate models/symdef/model.tar.gz data/test.json - -include-package dygie - -output-file models/symdef/metrics\_test.json''
    \item To compare results to TaDDEx, use the script \url{anonymous} to convert the output into our format.
\end{enumerate}

\section{GPT3 Experimental Setup}\label{appendix:gpt3}
We generated GPT3 baselines using text completion with text-davinci-003 in a question-answer format. We prepared the prompts by concatenating 
\begin{quotation}
Question: given the following sentence,
\end{quotation}
with the sample sentence, replacing each of $N$ symbols in the sentence with SYMBOL1, SYMBOL2, \dots, SYMBOLN, and appending one of the following based on the number of symbols to the end: \begin{quotation}
what are the definitions, if any, of SYMBOL1, SYMBOL2, \dots and SYMBOLN? Answer:
\end{quotation}

For example, the sentence
\begin{quotation}
Each word SYMBOL is embedded into a SYMBOL -dimensional word vector SYMBOL , where SYMBOL is a ( learned ) word embedding matrix , SYMBOL is the vocabulary size , and notation SYMBOL denotes the SYMBOL -th column of matrix SYMBOL .
\end{quotation}
would be transformed into the following prompt:
\begin{quotation}
Question: given the following sentence, "Each word SYMBOL1 is embedded into a SYMBOL2 -dimensional word vector SYMBOL3 , where SYMBOL4 is a ( learned ) word embedding matrix , SYMBOL5 is the vocabulary size , and notation SYMBOL6 denotes the SYMBOL7 -th column of matrix SYMBOL8 .", what are the definitions, if any, of SYMBOL1, SYMBOL2, SYMBOL3, SYMBOL4, SYMBOL5, SYMBOL6, SYMBOL7, and SYMBOL8? Answer:
\end{quotation}
To reduce the likelihood of GPT3 completing the prompt with text outside of the sample sentence, we set the temperature to 0.0.
\subsection{GPT3 One-Shot and Few-Shot Examples}
For the one-shot experiments, we prepended the following example to each prompt:
\begin{quotation}
Question: given the following sentence, ``It can be represented as: SYMBOL1 where SYMBOL2 is the bidirectional GRU, SYMBOL3 and SYMBOL4 denote respectively the forward and backward contextual state of the input text.'', what are definitions, if any, of SYMBOL1, SYMBOL2, SYMBOL3, and SYMBOL4? 

ANSWER: SYMBOL1 has no definition. SYMBOL2 is defined as bidirectional GRU. SYMBOL3 is defined as forward contextual state of the input text. SYMBOL4 is defined as backward contextual state of the input text.
\end{quotation}

For the few-shot experiments, we prepended four examples to each prompt:
\begin{quotation}
Question: given the following sentence, ``It can be represented as: SYMBOL1 where SYMBOL2 is the bidirectional GRU, SYMBOL3 and SYMBOL4 denote respectively the forward and backward contextual state of the input text.'', what are definitions, if any, of SYMBOL1, SYMBOL2, SYMBOL3, and SYMBOL4? 

ANSWER: SYMBOL1 has no definition. SYMBOL2 is defined as bidirectional GRU. SYMBOL3 is defined as forward contextual state of the input text. SYMBOL4 is defined as backward contextual state of the input text.

Question: given the following sentence, ``In general, gradient descent optimization schemes may fail to converge to the equilibrium by moving along the orbit trajectory among saddle points CITATION (CITATION).'', what is the definition, if any, of SYMBOL? 

ANSWER: There is no definition.

Question: given the following sentence, ``For each target emotion (i.e., intended emotion of generated sentences) we conducted an initial MANOVA, with human ratings of affect categories the DVs(dependent variables) and the affect strength parameter SYMBOL1 the IV (independent variable).'', what is the definition, if any, of SYMBOL1? 

ANSWER: SYMBOL1 is defined as affect strength parameter.

Question: given the following sentence, ``The CSG program in our example consists of two boolean combinations: union, SYMBOL1 and subtraction SYMBOL2 and two primitives: circles SYMBOL3 and rectangles SYMBOL4, specified by position SYMBOL5, radius SYMBOL6, width and height SYMBOL7, and rotation SYMBOL8.'', what are definitions, if any, of SYMBOL1, SYMBOL2, SYMBOL3, SYMBOL4, SYMBOL5, SYMBOL6, SYMBOL7, and SYMBOL8? 

ANSWER: SYMBOL1 is defined as union. SYMBOL2 is defined as subtraction. SYMBOL3 is defined as circles. SYMBOL4 is defined as rectangles. SYMBOL5 is defined as position. SYMBOL6 is defined as radius. SYMBOL7 is defined as height. SYMBOL8 is defined as rotation.
\end{quotation}

\subsection{GPT3 Post-Processing}
In order to fairly compare GPT3 to the other models in this study, its output must be reformatted into slot labels. Our post-processing script carries out the following steps: 
\begin{enumerate}
    \item Using regex, chunk the output according to the symbols in the sentence so that there is one snippet per symbol. The remaining steps are performed for each snippet.
    \item Using regex, detect whether the symbol definition is negative ("SYMBOL1 has no definition"). If so, assign all slot labels to ``O''.
    \item Words in the response that are not found in the sentence are deleted.
    \item Words in the response that only occur once in the sentence are automatically labeled as B-DEF (for the first word in the current response snippet) or I-DEF.
    \item Responses that have words with multiple occurrences in the sentence are printed and a human indicates which slot should be selected. 
\end{enumerate}

Multiple challenges arise in post-processing GPT3's output. Firstly, the responses often contain words that are not present in the sample text. This occurs when GPT3's output contains a meta-description of its own output (for example, it might print "It is not possible to say for certain what the definition of SYMBOL1 is. However, SYMBOL1 might be defined as..."). This text can be trimmed out using regex pattern matching. Additional text also occurs when GPT3 provides external information (for example, the presence of the phrase ``word embeddings'' may trigger GPT3 to provide general information about word embeddings rather than a definition for a symbol in the sentence). Our post-processing script deletes words in the output that are not present in the input, which can mitigate some instances of external information. However, sometimes the GPT3 response contains words that are in the sentence. If the gold label for such words is B-DEF or I-DEF, the GPT-3 score benefits despite the answer being inappropriate. If the gold label is ``O'', then the GPT-3 score suffers. Sometimes GPT3 copies a large span of text from the sentence as its answer. Sometimes this results in the GPT3 output containing a ``respectively'' construct, which is not useful since the purpose of our system is to resolve those constructs. An instance of this from the zero-shot experiments is as follows:
\begin{quote}
There are no specific definitions given for any of the symbols in the sentence. However, we can infer that SYMBOL1 refers to a pixel location inside of SYMBOL2, which is likely a box or other geometric shape. SYMBOL3 represents the projected box onto SYMBOL2, and SYMBOL4 is a 4-dimensional vector that encodes the projected box. SYMBOL5, SYMBOL6, SYMBOL7, and SYMBOL8 represent the distances between the current pixel location and the top, left, bottom, and right boundaries of the projected box, respectively.
\end{quote}
The post-processing difficulties, lack of consistency in responses, and lack of reliability in terms of truthfulness or appropriateness of responses make GPT3 inference difficult to use in this particular scientific document processing task.
\section{GPU Usage}
\label{appendix:gpu}
This section provides an estimation of GPU usage and the model sizes for \method and the baseline systems.

\textbf{Model sizes:}
\begin{itemize}
    \item TaDDEx and HEDDEx are based on RoBERTa-large, which contains 355 million parameters;
    \item DyGIE++ uses SciBERT \citep{beltagy-etal-2019-scibert} which contains 110 million parameters;
    \item SciIE uses ELMo \citep{peters-etal-2018-deep}, which contains 93.6 million parameters;
    \item and GPT3 contains 175 billion parameters.
\end{itemize}

Training and testing for TaDDEx, HEDDEx, DyGIE++, and SciIE was performed on a single NVIDIA RTX A6000 GPU. Using our training set as input, it takes approximately 3.5 hours to train TaDDEx, 3 hours to train HEDDEx, 6 hours to train DyGIE++, and 6 hours to train SciIE. These models were trained multiple times over the course of this study with an approximate GPU usage between 80 and 100 hours. 3,354 requests were made to GPT3's text-davinci-003 model, resulting in a total of 741,680 input and output tokens.

\end{document}